\documentclass[letterpaper]{article} 
\usepackage{aaai2026}  
\usepackage{times}  
\usepackage{helvet}  
\usepackage{courier}  
\usepackage[hyphens]{url}  
\usepackage{graphicx} 
\usepackage{natbib}  
\usepackage{caption} 
\usepackage{multirow} 
\usepackage{algorithm}
\usepackage{algorithmic}
\usepackage{amsmath}
\usepackage{amssymb}
\usepackage{xcolor}
\usepackage{arydshln} 
\usepackage{newfloat}
\usepackage{listings}
\DeclareCaptionStyle{ruled}{labelfont=normalfont,labelsep=colon,strut=off} 
\floatstyle{ruled}
\newfloat{listing}{tb}{lst}{}
\floatname{listing}{Listing}
\title{Robust Pedestrian Detection with Uncertain Modality}
\newcommand{\corrauthor}{\thanks{Corresponding author.}}
\author{
    Qian Bie\textsuperscript{\rm 1}\equalcontrib,
    Xiao Wang\textsuperscript{\rm 2}\equalcontrib,
    Bin Yang\textsuperscript{\rm 3}\corrauthor,
    Zhixi Yu\textsuperscript{\rm 1},
    Jun Chen\textsuperscript{\rm 3},
    Xin Xu\textsuperscript{\rm 1}\footnotemark[2]
}
\affiliations{
    \textsuperscript{\rm 1} School of Computer Science and Technology, Wuhan University of Science and Technology, Wuhan 430065 China \\
    \textsuperscript{\rm 2} Hubei Province Key Laboratory of Intelligent Information Processing and Real-time Industrial System, Wuhan University of Science and Technology, China\\
    \textsuperscript{\rm 3} National Engineering Research Center for Multimedia Software, School of Computer Science, Wuhan University, China \\
    bieqian@wust.edu.cn, wangxiao2021@wust.edu.cn, 
    yangbin\_cv@whu.edu.cn,\\
    yuzhixi@wust.edu.cn,
    chenj@whu.edu.cn,  xuxin@wust.edu.cn
}

\usepackage{bibentry}

\begin{document}

\maketitle

\begin{abstract}
Existing cross-modal pedestrian detection (CMPD) employs complementary information from RGB and \underline{t}hermal-\underline{i}nf\underline{r}ared (TIR) modalities to detect pedestrians in 24h-surveillance systems. RGB captures rich pedestrian details under daylight, while TIR excels at night. However, TIR focuses primarily on the person's silhouette, neglecting critical texture details essential for detection. 
While the \underline{n}ear-\underline{i}nf\underline{r}ared (NIR) captures texture under low-light conditions, which effectively alleviates performance issues of RGB and detail loss in TIR, thereby reducing missed detections. To this end, we construct a new \textbf{T}riplet \textbf{R}GB–\textbf{N}IR–\textbf{T}IR (TRNT) dataset, comprising 8,281 pixel-aligned image triplets, establishing a comprehensive foundation for algorithmic research. 
However, due to the variable nature of real-world scenarios, imaging devices may not always capture all three modalities simultaneously. This results in input data with unpredictable combinations of modal types, which challenge existing CMPD methods that fail to extract robust pedestrian information under arbitrary input combinations, leading to significant performance degradation.
To address these challenges, we propose the \textbf{A}daptive \textbf{U}ncertainty-aware Network (AUNet) for accurately discriminating modal availability and fully utilizing the available information under uncertain inputs. 
Specifically, we introduce Unified Modality Validation Refinement (UMVR), which includes an uncertainty-aware router to validate modal availability and a semantic refinement to ensure the reliability of information within the modality. 
Furthermore, we design a Modality-Aware Interaction (MAI) module to adaptively activate or deactivate its internal interaction mechanisms per UMVR output, enabling effective complementary information fusion from available modalities. 
AUNet enables accurate modality validation and robust inference without fixed modality pairings, facilitating the effective fusion of RGB, NIR, and TIR information across diverse inputs. 
Code: \url{https://github.com/medicineqian/AuNet.}
\end{abstract}

\begin{figure}[!t]
\centering
\includegraphics[scale=0.47]{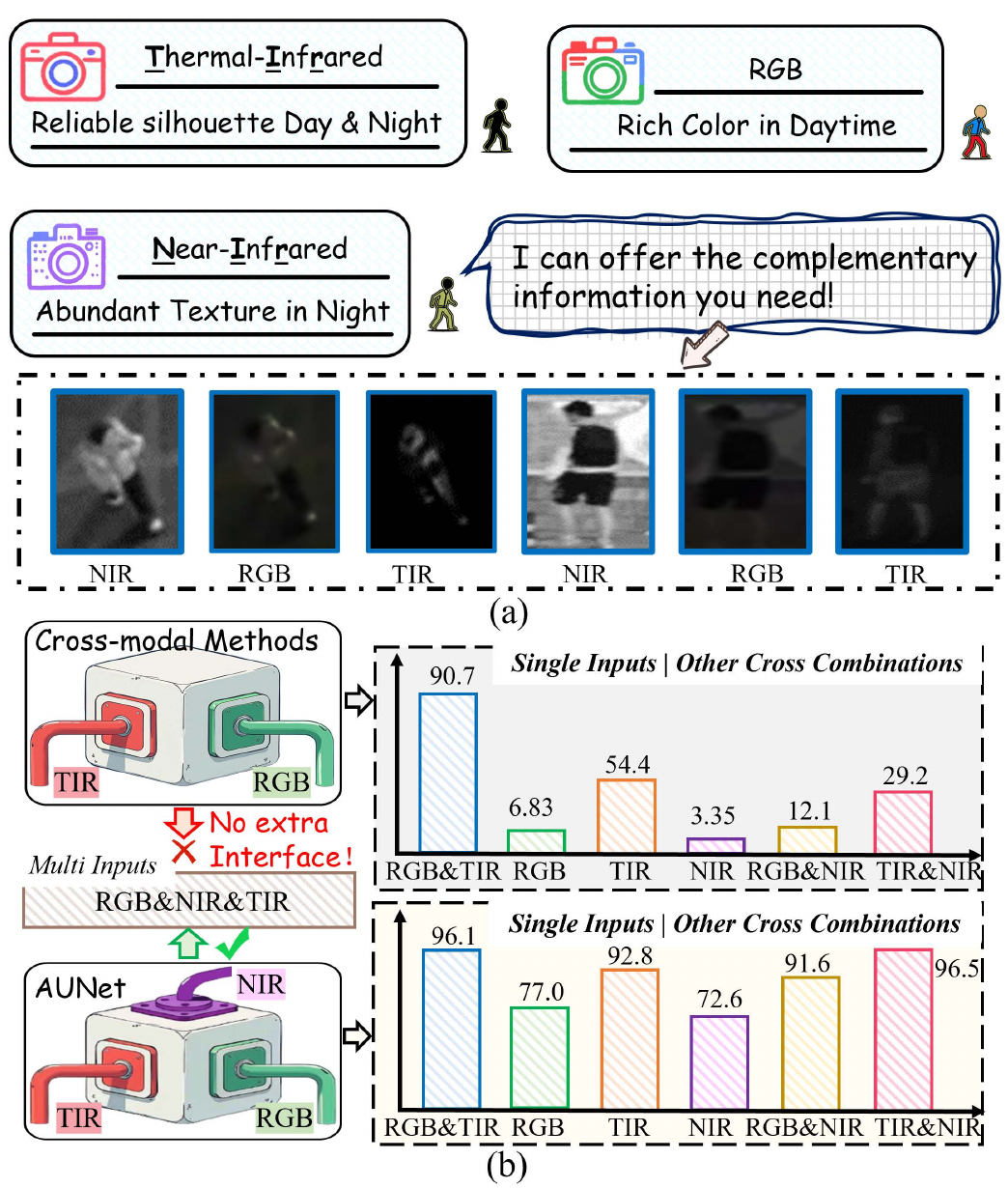}
\caption{Illustration of our idea. (a) demonstrates the complementary benefits of the NIR for RGB and TIR. (b) illustrates the performance degradation of existing CMPD methods under single-modal and other cross-modal inputs. we conduct evaluation using ICAFusion \cite{shen2024icafusion} and our AUNet trained on our DRNT.}
\label{Figure1}
\end{figure}
\section{Introduction}
\label{sec:intro}
Robust pedestrian detection is essential for round-the-clock applications such as autonomous driving \cite{kim2021mlpd,chen2022multimodal,zhouunbiased} and intelligent surveillance \cite{cao2023multimodal,chen2025deyolo,xu2025efficient,liu2024learning}, where cross-modal pedestrian detection (CMPD) leveraging complementary information from RGB and \underline{t}hermal-\underline{i}nf\underline{r}ared (TIR) data is a primary approach. Over the past decade, various CMPD methods have been proposed for RGB and TIR information fusion, ranging from direct feature concatenation \cite{liu2016multispectral,konig2017fully}, illumination-guided weighting \cite{li2019illumination,xie2023illumination}, gated mechanisms \cite{zheng2019gfd}, differential fusion \cite{zhou2020improving}, attention-based fusion \cite{zhang2021guided,shen2024icafusion}, frequency-domain fusion \cite{li2025fd2}, to causal inference mechanisms \cite{kim2024causal}.
However, TIR imaging primarily captures the coarse pedestrian's silhouette based on the principle of heat radiation, neglecting the vital texture details necessary for accurate detection, as depicted in Figure~\ref{Figure1} (a).
In contrast, the \underline{n}ear-\underline{i}nf\underline{r}ared (NIR) sensor is capable of capturing abundant person texture information, particularly under active nighttime illumination. Therefore, incorporating the complementary NIR (abundant texture at night) information for RGB (rich color and texture in daytime) and TIR (reliable silhouette day \& night) can effectively alleviate detail loss in TIR and target blurring in RGB images under low-light conditions, thereby reducing the rate of missed pedestrian detections.

\begin{figure*}[!t]
\centering
\includegraphics[scale=0.54]{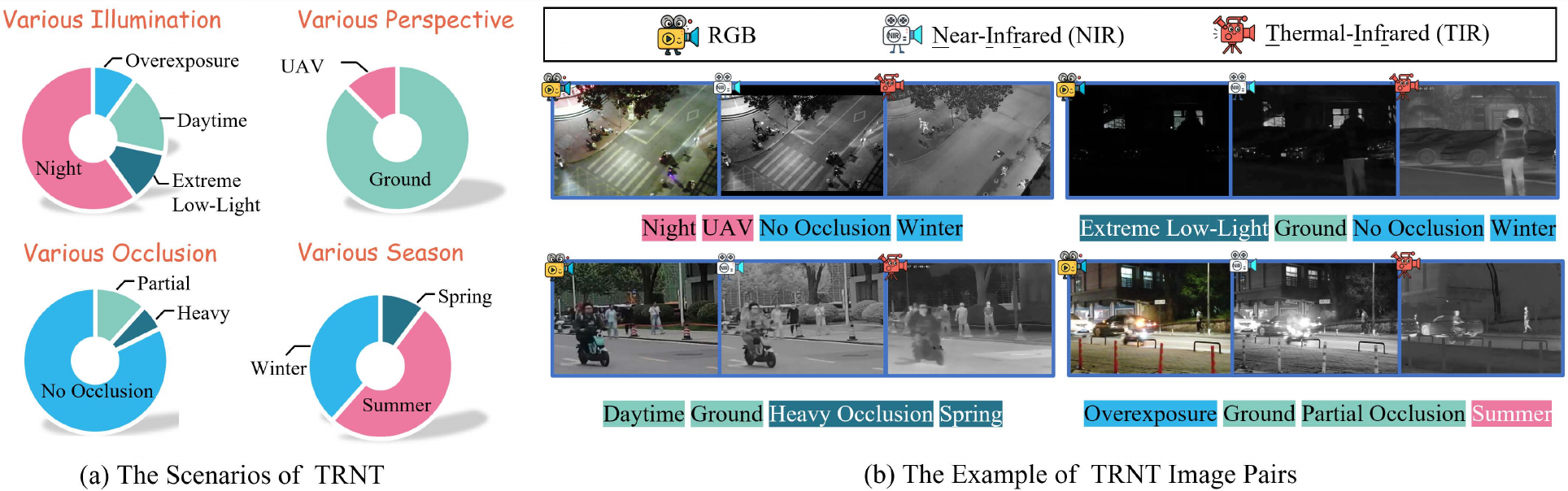}
\caption{ Data Diversity. The scenarios and examples of our benchmark TRNT. TRNT covers diverse scenarios with various illumination, perspective, occlusion, and season.}
\label{data_challenge}
\end{figure*}

To this end, we construct a novel comprehensive \textbf{T}riplet \textbf{R}GB–\textbf{N}IR–\textbf{T}IR (TRNT) dataset, captured by three co-collaborative sensors across diverse real-world scenarios. As illustrated in Figure~\ref{data_challenge} (a), it contains various detection-relevant factors, including variations in illumination (day $\xrightarrow{}$ night), occlusion (partial $\xrightarrow{}$ heavy), weather conditions (winter $\xrightarrow{}$ summer), and complex backgrounds. Furthermore, TRNT consists of 7,429 and 852 image triplets captured from the ground view and Unmanned Aerial Vehicle (UAV) perspective, respectively, accounting for dynamic camera angles in real-world deployment scenarios. TRNT provides a comprehensive and reliable experimental foundation for exploring the potential of these three modalities.

However, due to the variable factors in real-world scenarios, imaging devices may not always capture RGB, NIR, and TIR images simultaneously. This results in input data with various and unpredictable modal types and quantities. Existing CMPD methods are typically designed for fixed RGB-TIR input pairs, thus failing to extract reliable pedestrian information under single-modal and other cross-modal inputs, which dramatically decreases their detection performance, as demonstrated in Figure~\ref{Figure1} (b). Besides, they are limited to two branches and cannot provide additional interfaces for the NIR modality. Although completing missing modality data is an intuitive solution, it requires additional generative modules and heavily relies on the availability of fully complete three-modality data during training.

To address these challenges, we propose a novel \textbf{A}daptive \textbf{U}ncertainty-aware Network (AUNet), designed for robust pedestrian detection under arbitrary input combinations of RGB, NIR, and TIR data. Specifically, to accurately validate the availability of input modalities, we implement a Unified Modality Validation Refinement (UMVR) with two components: a lightweight uncertainty-aware router net to validate the availability of each modality, with its output yielding a binary decision value (0 or 1). 
Motivated by the success of CLIP's powerful global semantic extraction capabilities, we design a CLIP-driven semantic refinement that allows semantic priors extracted by CLIP's image encoders to serve as guiding signals, discriminating critical pedestrian features from cluttered backgrounds to ensure the reliability of information within each modality. 
To fully leverage the information of the available modalities, we design a Modality-Aware Interaction (MAI) module that adaptively acquires the fusion strategy and dynamically activates or deactivates its internal interaction mechanisms according to the availability value obtained by UMVR, effectively integrating complementary information from different modalities.
For thorough examination, we conduct extensive experiments on both our TRNT and the LLVIP \cite{jia2021llvip} to demonstrate the effectiveness and robustness of our AUNet under various uncertain input combinations. In summary, we launch a new benchmark for multi-modality pedestrian detection and a new baseline for uncertain-modal pedestrian detection. The main contributions can be summarized as follows:
\begin{itemize}
    \item We construct a multi-scenario, multi-modality \textbf{T}riplet \textbf{R}GB–\textbf{N}IR–\textbf{T}IR dataset called TRNT, comprising 8281 precisely aligned RGB-NIR-TIR image triplets simultaneously captured by triple co-aligned sensors in diverse real-world environments.
    \item We propose an Adaptive Uncertainty-aware Network (AUNet) designed to tackle the challenge of uncertain modal inputs, which incorporates a Unified Modality Validation Refinement (UMVR) and a Modality-Aware Interaction (MAI) module to accurately validate modal availability and effectively exploit complementary information from available modalities, respectively.
    \item We conduct a comprehensive evaluation of various state-of-the-art methods, along with detailed experimental analysis on both our TRNT and the LLVIP cross-modal dataset to demonstrate the effectiveness and robustness of our AUNet under arbitrary input combinations.
\end{itemize}
\section{Related Work}
\label{sec:formatting}
\subsection{Cross-modal Pedestrian Detection Datasets}
To overcome the illumination constraints inherent in RGB-based pedestrian detection, Hwang et al. \cite{hwang2015multispectral} pioneer cross-modal pedestrian detection by introducing the KAIST dataset, which aims to leverage complementary information from RGB and TIR data for accurate detection. CVC-14 \cite{gonzalez2016pedestrian} and FLIR \cite{c:01} are constructed to complement KAIST by motion-blurred scenarios relevant to autonomous driving. Jia et al. \cite{jia2021llvip} introduce LLVIP, a high-resolution RGB–TIR dataset designed for pedestrian detection in low-light conditions. Liu et al. \cite{liu2022target} propose the M3FD dataset, which incorporates adverse weather conditions such as rain and fog. However, these datasets overlook the prevalent NIR modality used in night vision systems, which can provide rich pedestrian texture details complementary to both RGB and TIR, thereby significantly limiting the potential for multi-modal synergy.
\subsection{Cross-modal Pedestrian Detection Methods}
 Liu et al. \cite{liu2016multispectral} craft four sophisticated convolutional neural network architectures and demonstrate that the halfway fusion model exhibits superior detection synergy.
 Illumination-aware Faster R-CNN \cite{li2019illumination} dynamically weights RGB and TIR sub-networks via illumination-based gating to optimize multispectral fusion. 
 Yang et al. \cite{yang2022baanet} design a BAANet with bi-directional adaptive attention gates, progressively distilling modality-specific noise and recalibrating features via illumination-guided interactions. Shen et al. \cite{shen2024icafusion} introduce a framework with dual cross-attention transformers to overcome the limitations of CNNs in local-range feature interaction. Fusion-mamba \cite{dong2024fusion} introduces Mamba for cross-modal fusion, designs state space channel swapping and dual state space fusion modules to reduce modality disparities. FD$^2$Net \cite{li2025fd2} extracts RGB and TIR information through high-frequency units and low-frequency units, respectively, and combines them with a parameter-independent fusion strategy for further enhancing feature representation. However, these methods primarily focus on exploring the complementary correlations between fixed TIR and RGB images, neglecting variable input modality types and counts, which significantly limits their applicability and generalizability.
\begin{table*}[!t]
\centering
\small
\begin{tabular}{lcccccccc}
\hline \multirow{2}{*}{Datasets} & \multirow{2}{*}{Number of images} & \multicolumn{3}{c}{Modality} & \multirow{2}{*}{Camera angle} & \multicolumn{2}{c}{Illumination condition} & \multirow{2}{*}{Pedestrian} \\
\cline { 3 - 5} \cline{7-8} & &\multicolumn{1}{c}{ RGB } & TIR & NIR & & Day &Night\\
\hline TNO & $261 \times 2$ & $\checkmark$ & $\checkmark$ & $\times$ & Shot on the ground & few & $\checkmark$ & few \\
 INO & $2100 \times 2$ & $\checkmark$ & $\checkmark$ & $\times$ & surveillance & $\checkmark$ & $\checkmark$ & few \\
 OSU & $285 \times 2$ & $\checkmark$ & $\checkmark$ & $\times$ & surveillance & $\times$ & $\checkmark$ & $\checkmark$ \\
KAIST & $4750 \times 2$ & $\checkmark$ & $\checkmark$ & $\times$ & driving & $\checkmark$ & $\checkmark$ & $\checkmark$ \\
CVC-14 & $849 \times 2$ & $\checkmark$ & $\checkmark$ & $\times$ & driving & $\checkmark$ & $\checkmark$ & $\checkmark$ \\
 FILR & $5258 \times 2$ & $\checkmark$ & $\checkmark$ & $\times$ & driving & $\checkmark$ & $\checkmark$ & $\checkmark$ \\
LLVIP & $15488 \times 2$ & $\checkmark$ & $\checkmark$ & $\times$ & surveillance & $\checkmark$ & few & $\checkmark$ \\
DRNT(ours) & $8281 \times 3$ & $\checkmark$ & $\checkmark$ & $\checkmark$ & Shot on the ground, UAV & $\checkmark$ & $\checkmark$ & $\checkmark$ \\
\hline
\end{tabular}
\caption{A comparative analysis of the TRNT dataset with existing cross-modal pedestrian detection datasets. The number of images represents the dataset's selection rate of one frame per second. TRNT comprises 8281 RGB-NIR-TIR image triplets (24,843 images). Moreover, the TRNT dataset ensures that each image contains at least one or more pedestrians.}
\label{tab:my_label}
\end{table*}
\section{TRNT Dataset}
To effectively utilize the complementary advantages of \underline{n}ear-\underline{i}nf\underline{r}ared (NIR), we construct a novel dataset, \textbf{T}riplet \textbf{R}GB–\textbf{N}IR–\textbf{T}IR (TRNT), encompassing all three modalities and supporting seven possible input combinations.
\subsubsection{Data Acquisition}
Data collection employs two camera platforms: a customized triplicated Pan-Tilt-Zoom (PTZ) to capture images from the viewpoint of a shot on the ground, and a purchased Matrice 350 Real-Time Kinematic (RTK) drone device to capture images from the UAV viewpoint. Both platforms are equipped with synchronized RGB, NIR, and TIR sensors to ensure simultaneous tri-modal image pair captures. According to the KAIST \cite{hwang2015multispectral} dataset protocol, raw TIR images are converted to grayscale to obtain the final TIR images. 
\subsubsection{Data Description}
TRNT comprises 8281 rigorously aligned RGB-NIR-TIR image triplets captured from different camera angles with diverse illumination conditions and various occlusions. The division of the TRNT dataset into training and testing follows CVC-14, comprising 7026 image pairs in the training subset and 1255 image pairs in the test set. The bounding boxes are annotated accurately with a resolution of 640 $\times$ 480 for each image. Crucially, every image contains at least one pedestrian. As shown in Table \ref{tab:my_label}, compared with the existing cross-modal pedestrian detection datasets, TRNT has the following main advantages:
\begin{itemize}
\item \textbf{Novel Modality Integration.} TRNT pioneers the inclusion of \underline{n}ear-\underline{i}nf\underline{r}ared (NIR) alongside RGB and \underline{t}hermal-\underline{i}nf\underline{r}ared (TIR) in pedestrian detection. NIR sensors are widely available in modern surveillance systems but overlooked in prior cross-modal pedestrian detection research. The inclusion of all three common modalities significantly expands the diversity of possible input combinations compared to cross-modal datasets.
\item \textbf{Dual-Perspective Coverage.} TRNT is the first multi-modal dataset that integrates both ground-views (mimicking vehicle and robotic perspectives) and high-altitude UAV perspectives ($\ge$50m altitude), particularly the UAV perspective that has never been present in any of the previous cross-modal pedestrian detection datasets.
\item \textbf{Longitudinal Illumination and Seasonal transitions.} TRNT images span 14 continuous hours (8:00–22:00) covering diverse lighting conditions: natural illumination (day, night), extremely low illumination (dark environments without urban/car lights). Besides, TRNT images span three distinct seasons (winter, spring, and summer), introducing dynamic modality reliability shifts. For example, thermal radiation attenuation from pedestrians' upper bodies due to insulation by heavy winter clothing (please refer to the Figure~\ref{data_challenge} (b)).
\end{itemize}
\section{Methods}
\begin{figure*}[!t]
\centering
\includegraphics[scale=0.357]{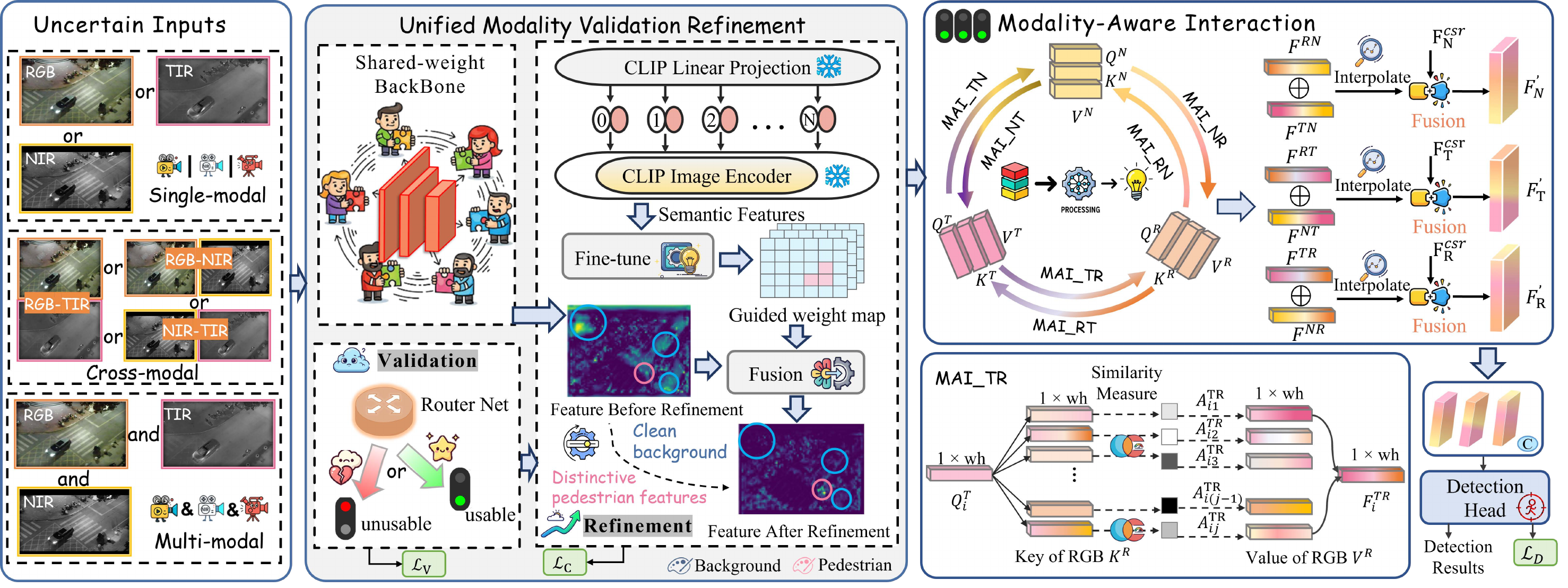}
\caption{Illustration of our AUNet, including two main components: 1) Unified Modality Validation Refinement (UMVR), which integrates an Uncertainty-aware Router (UAR) validation and CLIP-driven Semantic Refinement (CSR); 2) Modality-Aware Interaction (MAI), which effectively integrates complementary information from the available modalities. This figure depicts a scenario where all three modalities are available, corresponding to the UMVR output of [1, 1, 1].}
\vspace{-2mm}
\label{Figure2}
\end{figure*}

To dynamically handle arbitrary input combinations (single, dual, or triple modalities), we propose AUNet consisting of a Unified Modality Validation Refinement (UMVR) and a Modality-Aware Interaction (MAI) module to discriminate modal availability and exploit the information of the available modalities, respectively, as shown in Figure~\ref{Figure2}. 
\subsubsection{Shared-weight Feature Extraction}
Formally, let $\textit{I}_R \in \mathbb{R}^{H \times W \times 3} $, $\textit{I}_T \in \mathbb{R}^{H \times W \times 3} $, and $\textit{I}_N \in \mathbb{R}^{H \times W \times 3} $ be the input RGB, TIR, and NIR images, respectively. Unlike previous cross-modal works \cite{li2019illumination,chen2023attentive} that employ three independent backbones, we utilize a single weight-shared backbone $\mathcal{F}_{\theta}(\cdot)$ with shared parameters $\theta$ across all modalities, supporting flexible input configurations (single-, cross-, or multi-modal) without requiring architectural changes or retraining. If modality-specific backbones or partially shared backbones are employed instead, the AUNet architecture must be adjusted whenever the number of input modalities increases or decreases. Compared to separate backbone networks for each modality, a weight-shared backbone significantly decreases computational costs. The whole process can be represented as:
\begin{equation}\label{eqn-1}
  \begin{aligned}
    \textit{F}_X & = \mathcal{F}_{\theta}(\textit{I}_X), \quad X \in {R, T, N},
  \end{aligned}
\end{equation}
where RGB, TIR and NIR features are denoted as $ \textit{F}_R \in \mathbb{R}^{h \times w \times c}$, $ \textit{F}_T \in \mathbb{R}^{h \times w \times c}$, and $ \textit{F}_N \in \mathbb{R}^{h \times w \times c}$, respectively. Here, $\mathcal{F}_{\theta}(\cdot)$ can be any convolutional neural network. 
\subsubsection{Unified Modality Validation Refinement} To validate the availability of each input modality, we propose a lightweight uncertainty-aware router (UAR) to validate the availability of the input RGB, NIR, and TIR images. Specifically, UAR employs a parameterized multilayer perceptron (MLP) with sigmoid activations, with its output yielding three binary decision values $\mathbf{v} \in \{0, 1\}^3$. For example, when RGB, NIR, and TIR images are all available, the output decision of the router is $v=[1,1,1]$. To enforce consistency between UAR decisions and realistic availability $v^{'}\in \{0, 1\}^3 $ of input images, we introduce availability validation loss $\mathcal{L}_\text{V}$ applied per modality:
\begin{equation}\label{eqn-2}
  \begin{aligned}
    \mathcal{L}_\text{V} = \sum_{x \in X} \left[ -v_{x}^{\prime} \log v_{x} - (1 - v_{x}^{\prime}) \log (1 - v_{x}) \right]
  \end{aligned}
\end{equation}
While UAR can determine the availability of each modality, it doesn't guarantee that all its information will benefit detection. For instance, as shown in Figure~\ref{Figure2}, given nighttime RGB inputs, unrefined feature maps exhibit misguided attention toward artifacts, such as reflective pavement surfaces or luminous streetlights, rather than pedestrians. To mitigate this interference of intrinsic modality noise with the aggregation of available modal information (UAR output value is 1), we introduce a CLIP-driven specific refinement (CSR) module. The CSR module leverages semantic priors extracted by CLIP's image encoders as guiding signals to discriminate critical pedestrian features from background noise across three modalities. Specifically, CLIP-derived weighting maps $M$ are generated to suppress irrelevant background regions while enhancing pedestrian-aware features. Since the weights of CLIP are fixed, $M$ is fine-tuned to adapt to the pedestrian detection task. The process for RGB is formulated as:
\begin{equation}\label{eqn-2-1}
  \begin{aligned}
     M_R & = \delta(conv(conv(\mathcal{E}_\text{CLIP}(I_R)))),
  \end{aligned}
\end{equation}
where $\delta$ denotes the sigmoid activation function, $conv$ represents a $1 \times 1$ convolution operation, and $\mathcal{E}_\text{CLIP}(\cdot)$ denotes CLIP's image encoder. The refined feature $F_R^{csr}$ is then obtained via:
\begin{equation}\label{eqn-3}
  \begin{aligned}
    F^{csr}_R & = F_R \oplus F_R \odot M_R,
  \end{aligned}
\end{equation}
where $\oplus$ and $\odot$ represent element-wise sum and element-wise multiplication, respectively. Following the same methodology, we can derive the output refined maps for TIR and NIR, denoted as $F^{csr}_T$ and $F^{csr}_N$, respectively. To ensure the maps ($M_R$,$M_N$,$M_T$) generated by CSR precisely localize regions where features need to be enhanced or diminished, ground-truth pedestrian distribution maps are utilized for supervision. The process for acquiring realistic pedestrian distribution maps is formulated as:
\begin{equation}\label{eqn-4}
  \begin{aligned}
    F^{csr}_{truth} & = \frac{(x-cx)^2}{a^2} + \frac{(y-cy)^2}{b^2} < 1 + \epsilon ,
  \end{aligned}
\end{equation}
where $x$ and $y$ are acquired by the mesh grid function. $(cx,cy)$, $a$, and $b$ correspond to the coordinates of the center point and half the width and height of a bounding box, respectively. $\epsilon$ is set to 0.0001. The CSR is optimized by minimizing the contrastive loss between the $F^{csr}$ and the ground-truth pedestrian distribution maps. Taking the contrastive loss in the CSR module of RGB as an illustrative example, the contrastive loss $\mathcal{L}_{CR}$ is formulated as:
\begin{equation}\label{eqn-5}
\begin{split}
  \begin{aligned}
    \mathcal{L}_{CR} = \frac{1}{N}\sum_{i=1}^{N}(1-\frac{2|F^{csr}_R \cap F^{csr}_{truth}| + 1}{|F^{csr}_R| + |F^{csr}_{truth}| + 1}) \\
    = \frac{1}{N}\sum_{i=1}^{N}(1-\frac{\left(2 \times \sum_{p=1}^{h \times w} F^{csr}_{R\_p}F^{csr}_{truth\_p} \right)+ 1}{\left(\sum_{p=1}^{h \times w}F^{csr}_{R\_p}\right)+\left(\sum_{p=1}^{h \times w} F^{csr}_{truth\_p}\right)+ 1})&,
  \end{aligned}
\end{split}
\end{equation}
where $F^{csr}_{truth\_p}$ and $F^{csr}_{R\_p}$ denote the elements of $F^{csr}_{truth}$ and $F^{csr}_R$ at the location $p$, respectively. $N$ denotes the number of samples in a mini-batch. The total contrastive loss $\mathcal{L}_C$ is formulated as:
\begin{equation}\label{eqn-6}
  \begin{aligned}
    \mathcal{L}_C & = W_R\mathcal{L}_{CR} +  W_T\mathcal{L}_{CT} +  W_N\mathcal{L}_{CN},
  \end{aligned}
\end{equation}
where $W_R$, $W_T$, and $W_N$ represent the learnable fusion weight parameters, initially set to 1.0.
\begin{table*}[!t]
\renewcommand{\arraystretch}{1.25}
\centering
\small
\begin{tabular}{lc:c:ccc:c:ccc:c:c}
\hline \multirow{2}{*}{Methods} & \multicolumn{3}{c}{VGG16} & & \multicolumn{3}{c}{ResNet50} & &\multicolumn{3}{c}{CSPDarknet53} \\
\cline { 2 - 4} \cline{6-8} \cline{10-12} &\multicolumn{1}{c}{AP$_{50}\uparrow$} & AP$_{75}\uparrow$ & mAP$\uparrow$ && AP$_{50}\uparrow$  & AP$_{75}\uparrow$ & mAP$\uparrow$ && AP$_{50}\uparrow$  & AP$_{75}\uparrow$ & mAP$\uparrow$\\
\hline RGB & 73.91 & 23.70 & 31.38 & & 78.03 & 26.23 & 34.33 & & 77.00 & 22.03 & 32.00 \\
NIR & 68.94 & 18.89 & 27.99 & &72.89 & 19.79 & 29.81 & &72.61 & 20.21 & 29.83 \\
TIR & 90.78 & 46.48 & 47.80 & & 92.38 & 53.60 & 51.93 & &92.81 & 52.22 & 51.36\\
(RGB-NIR)\_AUNet & 91.73 & 45.79 & 46.97 & & 91.48 & 49.97 & 50.06 & & 91.63 & 49.64 & 50.29\\
(NIR-TIR)\_AUNet & 95.24 & 62.27 & 57.20 & & 96.50 & 64.85 & 57.48 & & 96.51 & 64.92 & 57.74\\
(RGB-TIR)\_AUNet & 95.93 & 63.24 & 57.63 & & 96.86 & 66.33 & 59.12 & & 96.10 & 66.27 & 58.32\\
(RGB-NIR-TIR)\_AUNet & \textbf{96.63} & \textbf{65.49} & \textbf{58.24} & & \textbf{97.03} & \textbf{67.30} & \textbf{59.82} & &\textbf{97.07} & \textbf{67.78} & \textbf{60.08}\\
\hline
\end{tabular}
\caption{ Experimental comparison of the different modalities and backbones on TRNT (in \%). It is obvious that AUNet with CSPDarknet53 achieves the best performance compared to the other backbones.}
\label{tab:backbone}
\end{table*}
\subsubsection{Modality-Aware Interaction}
To adaptively acquire the fusion strategy per UMVR output and effectively integrate complementary information from available modalities, we introduce the Modality-Aware Interaction (MAI) module with a diversified information interaction mechanism. As illustrated at the right of Figure~\ref{Figure2}, the MAI module contains six predefined and independent cross-modal information interaction mechanisms called MAI\_TN, MAI\_NT, MAI\_TR, MAI\_RT, MAI\_NR, and MAI\_RN, respectively. When RGB, TIR, and NIR are all available, MAI activates all cross-modal interaction mechanisms to capture complementary information among the three modalities. When only two modalities are available, for example, RGB and TIR, MAI activates the MAI\_TR and MAI\_RT. To elaborate on the process of the interaction mechanism, 
we take the MAI\_TR interaction as an example without loss of generality. As shown in the bottom right corner of Figure~\ref{Figure2}, the feature from $F^{csr}_T$ is passed into a pooling layer, a linear, and layernorm transformation to generate a query matrix $Q^{T} = \{Q^{T}_1,\cdots, Q^{T}_i,\cdots, Q^{T}_{L}\}$. Similarly, the feature from $F^{csr}_R$ is passed into two pooling layers, two linear, and layernorm transformations to generate a key matrix $K^{R} = \{K^{R}_1,\cdots, K^{R}_j,\cdots, K^{R}_{L}\}$ and a value matrix $V^{R} = \{V^{R}_1,\cdots, V^{R}_j,\cdots, V^{R}_{L}\}$. $L$ denotes the number of queries, keys, and values in MAI\_TR. The i-th element of $F^{TR}$ is calculated as follows:  
\begin{equation}\label{eqn-7}
  \begin{aligned}
    F^{TR}_{i} & = \sum_{j=1}^{L}V^{R}_{j} \cdot A^{TR}_{ij},
  \end{aligned}
\end{equation}
\begin{equation}\label{eqn-8}
  \begin{aligned}
    A^{TR}_{ij} & = \frac{\exp(s_{ij})}{\sum_{j=1}^{L}\exp(s_{ij})}, \ \ \ \ s_{ij}&=\frac{Q^{T}_{i} (K^{R}_{j})^\top}{\sqrt{c}},
  \end{aligned}
\end{equation}
where $(\cdot)^{\top}$ means the matrix transposition. Here, $A^{TR}_i = \{A^{TR}_{i1},\cdots, A^{TR}_{ij},\cdots, A^{TR}_{iL}\} \in R^{1 \times L}$ represents the relevancy of the correlation between $Q^{T}_i$ and $K^{R}$. $ F^{TR} = \{F^{TR}_{i}\}^L_{i=1}$. We can acquire the output $F^{RN}$, $F^{TN}$, $F^{NR}$, $F^{TR}$, $F^{RT}$ and $F^{NT}$ by six independent information interactions. Taking $F^{RN}$ and $F^{TN}$ as examples, they are the complement features of RGB to NIR and TIR to NIR, respectively. $F^{RN}$ and $F^{TN}$ are summed with learnable weights with $F^{csr}_N$. The process for acquiring final NIR maps is expressed as:
\begin{equation}\label{eqn-9}
  \begin{aligned}
   F^{\prime}_N & = \mathcal{I}(W_{RN}F^{RN} \oplus W_{TN}F^{TN}) \oplus F^{csr}_N,
  \end{aligned}
\end{equation}
where $\mathcal{I}(\cdot)$ means interpolation. The output feature maps of MAI are reshaped to output feature maps of CSR. Here, $W_{RN}$ and $W_{TN}$ are two different learnable weights. Following this analogy, we can derive $F^{\prime}_T$ and $F^{\prime}_R$. Finally, the $F^{\prime}_N$, $F^{\prime}_R$, and $F^{\prime}_T$ are concatenated by two $1\times1$ convolution blocks, followed by batch normalizations and RELU activations, to acquire the final fused map called $F^{\prime}_m$.
\subsubsection{Objective Function}
As illustrated in Figure~\ref{Figure2}, we utilize a multi-scale detection head to detect pedestrians of different sizes. Our objective function comprises three components: loss for UAR ($\mathcal{L}_V$), loss for CSR ($\mathcal{L}_C$), and loss for the detection module ($\mathcal{L}_D$) that contains classification and localization loss of our one-stage AUNet. 
\begin{equation}\label{eqn-10}
  \begin{aligned}
    \mathcal{L}  =  \mathcal{L}_D + \lambda_1 \mathcal{L}_V + \lambda_2 \mathcal{L}_C , \mathcal{L}_D = \mathcal{L}_{cls} + \mathcal{L}_{loc},
  \end{aligned}
\end{equation}
where $\lambda_1$ and $\lambda_2$ denote the learnable weight parameters, initially set to 0.5 in the experiments.
\section{Experiments}
\begin{table}[!t]
\centering
 \small
\begin{tabular}{lc:c:cc}
\hline \multirow{2}{*}{Methods} & \multicolumn{3}{c}{ TRNT } & \multirow{2}{*}{FPS(Hz)$\uparrow$ }\\
\cline { 2 - 4 } &\multicolumn{1}{c}{ AP$_{50}\uparrow$ } & \multicolumn{1}{c}{ AP$_{75}\uparrow$ } & \multicolumn{1}{c}{ mAP$\uparrow$}\\
\hline 
ProbEn & 86.29 & 50.64 & 41.23 & 6.97 \\
TINet & 96.60 & 64.10 & 56.47 & 10.40\\
INSANet & 50.40 & 17.50 & 13.80 & 42.36\\
ICAFusion & 95.32 & 64.19 & 56.68 & 32.62\\
DE-YOLO & 95.73 & 65.26 & 57.00 & 38.64\\
AUNet  & \textbf{97.07} & \textbf{67.78} & \textbf{60.08}& \textbf{60.99}\\
\hline
\end{tabular}
\caption{Experimental results of AUNet on TRNT comparing with state-of-the-art methods while handling multi-modality pedestrian detection data.}
\vspace{-2mm}
\label{sota_methods}
\end{table}
\subsection{Impact of Different Input Combinations}
To ascertain the contribution of NIR information and validate the robustness of AUNet under seven possible modality inputs, we conduct a comprehensive evaluation of our method across diverse modal combinations on three different backbones, including VGG16 \cite{simonyan2014very}, ResNet50 \cite{he2016deep}, CSPDarkNet53 \cite{bochkovskiy2020yolov4}. As shown in Table \ref{tab:backbone}, this outcome ascertains the information contribution for RGB and TIR. Specifically, the AP$_{75}$ achieved by RGB-NIR (49.64\%) with our AUNet is higher than the single RGB (22.03\%) and NIR (20.21\%) added together when utilizing CSPDarkNet53 as the backbone. Even if we change the backbone, the proposed AUNet with RGB-NIR-TIR inputs still outperforms other methods. Furthermore, it demonstrates AUNet’s capability to dynamically harness arbitrary subsets of RGB-NIR-TIR modalities, mitigating the negative impact of uncertainty inputs through Unified Modality Validation Refinement and Modality-Aware Interaction, which greatly increases its applicability and generalizability. In light of its ranking performance, particularly concerning the mean Average Precision (mAP) metric, we select CSPDarkNet53 as our default backbone architecture. 
\begin{table*}[!t]
\renewcommand{\arraystretch}{1.25}
\centering
\small
\begin{tabular}{lc:c:ccc:c:ccc:c:c}
\hline \multirow{2}{*}{Methods} & \multicolumn{3}{c}{VGG16} & & \multicolumn{3}{c}{ResNet50} & &\multicolumn{3}{c}{CSPDarknet53} \\
\cline { 2 - 4} \cline{6-8} \cline{10-12} &\multicolumn{1}{c}{AP$_{50}\uparrow$} & AP$_{75}\uparrow$ & mAP$\uparrow$ && AP$_{50}\uparrow$  & AP$_{75}\uparrow$ & mAP$\uparrow$ && AP$_{50}\uparrow$  & AP$_{75}\uparrow$ & mAP$\uparrow$\\
\hline Baseline & 95.68 & 61.73 & 55.33 & & 96.02 & 62.10 & 56.29 & & 96.10 & 62.16 & 56.06 \\
+MAI & 96.20 & 62.98 & 56.68 & & 96.50 & 63.62 & 57.74 & & 96.85 & 63.95 & 57.94 \\
+MAI+CSR (AUNet) & \textbf{96.63} & \textbf{65.49} & \textbf{58.24} & & \textbf{97.03} & \textbf{67.30} & \textbf{59.82} & &\textbf{97.07} & \textbf{67.78} & \textbf{60.08}\\
\hline
\end{tabular}
\caption{An ablation study is conducted on the CSR and the MAI within the TRNT framework, utilizing various backbones. The baseline model refers to the direct concatenation of RGB-NIR-TIR modality features without CSR or MAI. The progressive enhancement in detection performance is attributed to the sequential integration of the key modules, one at a time.}
\vspace{-2mm}
\label{tab:ablation}
\end{table*}
\begin{figure}[!t]
\centering
\includegraphics[scale=0.098]{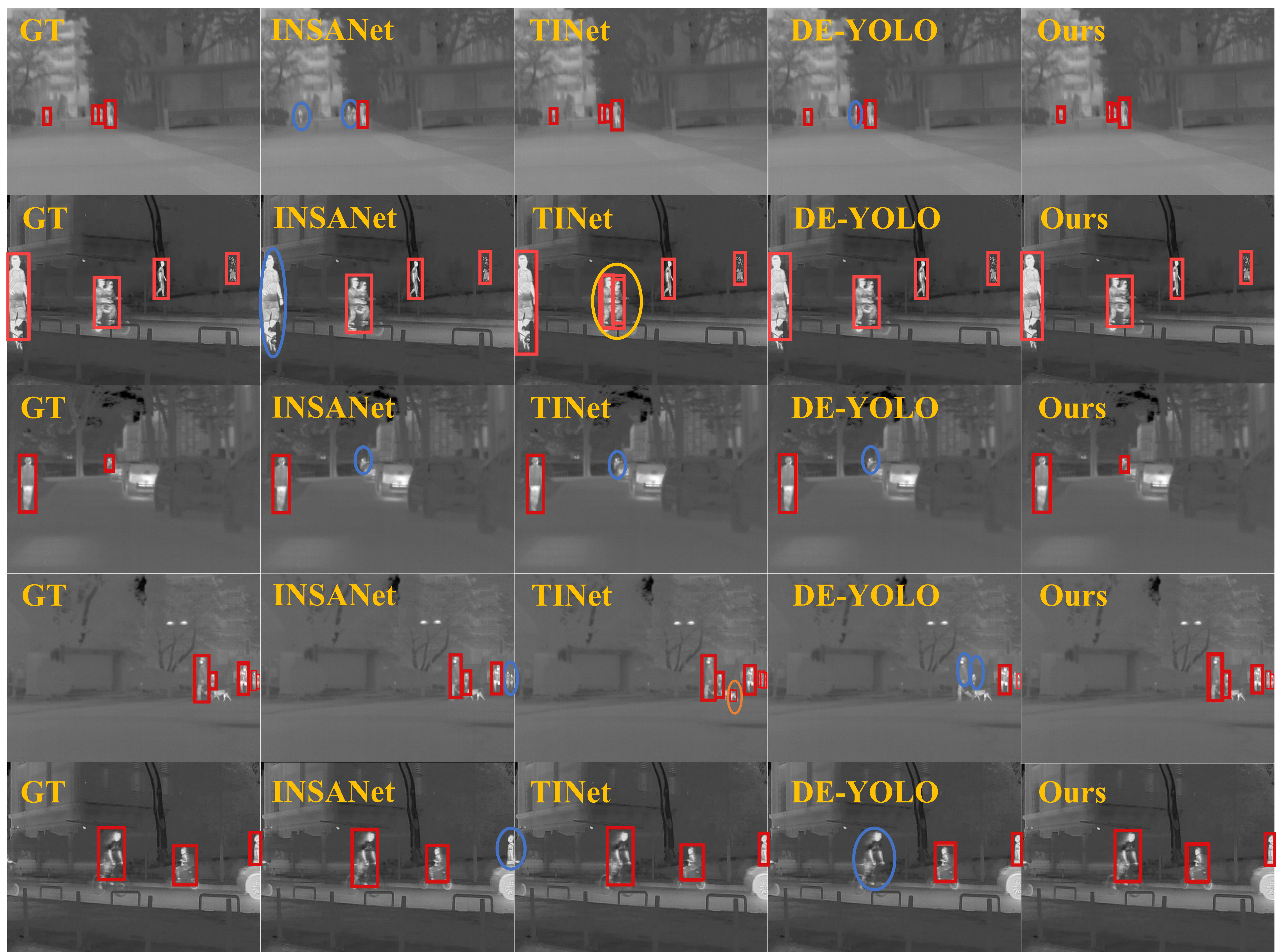}
\caption{Qualitative Comparison of INSANet, TINet, DE-YOLO, and our AUNet on the TRNT dataset. Red boxes indicate detection results. Blue circles highlight pedestrians missed. Orange circles mark the location of an incorrect detection. Yellow circles represent overlapping multiple detection boxes for the same instance.}
\label{sota_figure}
\end{figure}
\begin{table}[!t]
\renewcommand{\arraystretch}{1.25}
\centering
\small
\begin{tabular}{lcc:c:c}
\hline \multirow{2}{*}{Methods}&\multirow{2}{*}{Venue} & \multicolumn{3}{c}{ LLVIP } \\
\cline { 3-5} &&\multicolumn{1}{c}{ AP$_{50}\uparrow$ } & \multicolumn{1}{c}{ AP$_{75}\uparrow$ } & \multicolumn{1}{c}{ mAP$\uparrow$} \\
\hline Halfway Fusion & \textcolor{black}{BMVC'16} & 91.4 & 60.1 & 55.1 \\
GAFF & \textcolor{black}{WACV'21} & 94.0 & 60.2 & 55.8 \\
ProEn & \textcolor{black}{ECCV'22}& 93.4 & 50.2 & 51.5 \\
CMDet &\textcolor{black}{TCSVT'22}& 96.3 & - & -\\
CAMF &\textcolor{black}{TMM'23}& 89.0 & - & 55.6\\
MetaFusion &\textcolor{black}{CVPR'23} & 91.0 & - & 56.9 \\
CALNet &\textcolor{black}{MM'23} & - & - & 63.9\\
CSAA &\textcolor{black}{CVPR'23} & 94.3 & 66.6 & 59.2 \\
ICAFusion &\textcolor{black}{PR'24} & 96.3 & 71.7 & 62.3\\
Fusion-mamba &\textcolor{black}{Arxiv'24} & 96.5 & - & 62.8\\
FD$^2$Net &\textcolor{black}{AAAI'24} & 96.2 & 70.0 & -\\
AUNet &\textcolor{black}{---} &\textbf{97.2} & $\mathbf{73.5}$ & $\mathbf{64.8}$\\
\hline
\end{tabular}
\caption{Experimental results of AUNet on LLVIP compared with state-of-the-art methods while handling RGB-TIR cross-modality pedestrian detection data. Where `-' means the corresponding result is not available.}
\label{tab:llvip_sota}
\end{table}
\subsection{Comparison with State-of-the-art Methods}
Since it is the first work of multi-modality pedestrian detection, we extend five state-of-the-art cross-modality pedestrian detection methods (ProbEn \cite{chen2022multimodal}, TINet \cite{zhang2023illumination}, INSANet \cite{lee2024insanet}, ICAFusion \cite{shen2024icafusion}, and DE-YOLO \cite{chen2025deyolo}) and fuse the deep features from the NIR modality with their RGB-TIR fused feature maps for comparison. 
As reported in Table \ref{sota_methods}, INSANet introduces an attention-based novel fusion network to capture global intra- and inter-information. TINet designs an Intra-modality and Inter-modality module and fuses the inter- and intra-modality features with illumination-guided feature weights for pedestrian detection. DE-YOLO aims to achieve the detection-center mutual enhancement of RGB-TIR by semantic-spatial cross-modality and bi-directional decoupled focus modules. ICAFusion and DE-YOLO achieve remarkable performance while handling uncertain-modality scenarios. However, they are still significantly inferior to our AUNet (mAP improves \textbf{3.08\%} compared to DE-YOLO). The suboptimal performance of INSANet (mAP 13.80\%) is due to its approach of fusing features processed through the INSA modules using fixed static weights rather than employing adaptive, learnable weights. As depicted in Figure~\ref{sota_figure}, INSANet and DE-YOLO exhibit more missed detections, while TINet tends to generate multiple high-confidence score boxes for one instance and is prone to false positives. Furthermore, we test the FPS of these methods and our AUNet on the RTX 3090 platform. Our AUNet runs at \textbf{60.99} Hz, which demonstrates that AUNet combines fast inference speed with robust detection performance. 

Furthermore, we conduct a comprehensive evaluation of AUNet with state-of-the-art CMPD methods on LLVIP to evaluate the performance of AUNet in handling RGB-TIR cross-modal pedestrian detection, as shown in Table \ref{tab:llvip_sota}. These detectors include Halfway Fusion \cite{liu2016multispectral}, GAFF \cite{zhang2021guided}, CFT \cite{qingyun2021cross}, ProEn \cite{chen2022multimodal}, CMDet \cite{sun2022drone},CAMF \cite{tang2023camf}, MetaFusion \cite{zhao2023metafusion}, CALNet \cite{he2023multispectral}, CSAA \cite{cao2023multimodal}, ICAFusion \cite{shen2024icafusion}, Fusion-mamba \cite{dong2024fusion} and FD$^2$Net \cite{li2025fd2}. Through this rigorous comparison, we aim to highlight the robustness and accuracy of AUNet (marking a \textbf{2.0\%} mAP enhancement compared to the FD$^2$Net), demonstrating its superior performance in the cross-modal pedestrian detection task.
\subsection{Ablation Studies}
We conduct an ablation study on the CSR and MAI of AUNet on TRNT, as reported in Table \ref{tab:ablation}. As expected, both MAI (mAP \textbf{$\uparrow$ +1.88\%} vs. 56.06\%) and UMVR (mAP \textbf{$\uparrow$ +2.14\%} vs. 57.94\%) enhance the performance of the baseline, which validates the effectiveness of each module in handling complex RGB-NIR-TIR multi-modal pedestrian detection scenarios. Our AUNet consistently outperforms the concatenation operation across all evaluated scenarios and metrics, appraising the superior effectiveness of the AUNet in multi-modal information fusion. Additionally, we provide a visual figure of the detection results for the ablation of modules. 
\section{Conclusion and Discussion}
To explore the potential of the near-infrared spectrum, which provides critical pedestrian texture information under low-light conditions, we first establish a spatially and temporally aligned TRNT dataset, breaking the fixed RGB-TIR constraint in existing cross-modal datasets. TRNT provides 8,281 rigorously aligned RGB-NIR-TIR image triplets that account for key challenges in pedestrian detection, including variations in illumination (day $\xrightarrow{}$ night), occlusion (partial $\xrightarrow{}$ heavy), weather conditions (spring $\xrightarrow{}$ winter), perspectives (ground $\xrightarrow{}$ UAV), and backgrounds (clean $\xrightarrow{}$ cluttered).
To achieve robust detection in real-life applications, where the input modal types and quantities are uncertain, we propose a novel Adaptive Uncertainty-aware Network (AUNet) to dynamically handle arbitrary input combinations of RGB, NIR, and TIR data, which is more flexible than existing CMPD methods. AUNet integrates a Unified Modality Validation Refinement module for reliable modal availability validation and a Modality-Aware Interaction (MAI) module for effectively exploiting complementary information from available modalities. Extensive experiments confirm AUNet's effectiveness and robust performance under uncertain inputs on both TRNT and LLVIP datasets. AUNet is the unique solution that supports arbitrary modality combinations, whereas existing methods require training a separate model for each input configuration.
\section{Acknowledgments}
This work was supported by the Natural Science Foundation of China (62302351, 62501428, and 62376201), Hubei Provincial Science \& Technology Talent Enterprise Services Program (2025DJB059), and Hubei Provincial Special Fund for Central-Guided Local S\&T Development (2025CSA017).

\bibliography{aaai2026}
\end{document}